\documentclass{article}

\usepackage{arxiv}

\usepackage[utf8]{inputenc} 
\usepackage[T1]{fontenc}    
\usepackage{hyperref}       
\usepackage{url}            
\usepackage{booktabs}       
\usepackage{amsfonts}       
\usepackage{nicefrac}       
\usepackage{microtype}      
\usepackage{lipsum}
\usepackage{graphicx}
\usepackage{listings}
\usepackage{CJK}
\usepackage{indentfirst}
\usepackage{amsmath}
\usepackage{fancyhdr}
\usepackage{tikz}
\graphicspath{ {./images/} }

\title{A Simple Voting Mechanism for Online Sexist Content Identification}

\author{
 Chao Feng \\
  University of Zurich\\
  Rämistrasse 71\\
  CH-8006 Zürich, Switzerland \\
  \texttt{chao.feng2@uzh.ch} \\
}

\begin{document}
\maketitle
\begin{abstract}
This paper presents the participation of the MiniTrue team in the EXIST 2021 Challenge on the sexism detection in social media task for English and Spanish. Our approach combines the language models with a simple voting mechanism for the sexist label prediction. For this, three BERT based models and a voting function are used. Experimental results show that our final model with the voting function has achieved the best results among our four models, which means that our voting mechanism brings an extra benefit to our system. Nevertheless, we also observe that our system is robust to data sources and languages.
\end{abstract}


\let\thefootnote\relax\footnotetext{Copyright \textcopyright\ 2020 for this paper by its authors. Use permitted under Creative Commons License Attribution 4.0 International (CC BY 4.0). IberLEF 2021, September 2021, Málaga, Spain.}

\section{Introduction}

Equality, openness, and freedom, as the foundation and pillar of the Internet spirit, should have made our cyberworld a better place. However, inequality, prejudice and bias against females still deeply implants in online social networks~\cite{ref_article1}. Sexist contents in social networks affect females’ life in both mental and physical sides, for instance the career equality, life quality, even the mental health~\cite{ref_proc2,ref_proc3}. So that, an automatic sexism identification system is urgently required, which could help to build a better cyberworld with equality, openness, and freedom.

This work is done as part of the EXIST 2021 (Sexism Identification in Social Networks) shared task~\cite{ref_proc3}, which is a classification task for online social media data in English and Spanish.  We are interested in the sub-task one called Sexism Identification. This sub-task is a binary classification problem, we aim to combine sentence level embeddings learned by deep-learning models like BERT~\cite{ref_bert} with a voting mechanism, to create a sexism identification system and help for fighting against online sexist. To reduce the noise of social media as well as the multilingual noise, we leverage various features of language models, and implement our system in three separate stages:
\begin{enumerate}
\item In order to solve the multilingual problem, we use three different pretrained language models in produce the sentence level embedding, including the original English version BERT~\cite{ref_bert}, the Multilingual BERT~\cite{ref_mbert}, and the Spanish version pretrained language model BETO~\cite{ref_beto}.

\item Three different models are built upon these different embeddings and inference the output label independently.

\item We use a simple voting mechanism to get the final label.
\end{enumerate}

The note is organized as follows. We present the description of our method in Section 2. Then we group the experimental results in Section 3 before discussing the perspectives and conclude in Section 4.

\section{System Architecture}

This section presents the architecture and the description of our system. Three basic models and the vote-inference system are described respectively in Subsection 2.1 and Subsection 2.2.

\subsection{Basic Models}
We develop three independent basic models for label prediction by combining language models with simple neural networks. As before mentioned, three language models are applied in our system, including BERT, Multilingual BERT, and BETO. 

\subsubsection{Basic Model One:}

This first basic model connect the language models with a feed forward network. In order to gain the multilingual repersentation for the input text data, our model uses two different pretrained language model, the BERT and the Multilingual BERT. After tokenization, the text data feed into these two language model respectively, and generate two sentence embeddings. A concentration layer is used to combine these two embedding together. After that, embedding data flow to a feed forward network for the final inference $I_1$.

\subsubsection{Basic Model Two:}

Research points out that the performance of Multilingual BERT model is not as good as the language specific language models~\cite{ref_beto}. On this count, we replace the Multilingual BERT by Spanish version language model, BETO. Similar with the first model, we also use the concentration layer to combine the English embedding with the Spanish embedding, and feed them into a inference network to get the final output $I_2$.

\subsubsection{Basic Model Three:}

To leverage the sequential information of the input text, we use a Bi-directional LSTM network to obtain the consecutive features after the language model. In this time, we just use a single multilingual BERT, and get the sequential embeddings for each input token. After the Bi-directional LSTM, we use a Sigmoid function to get the final inference $I_3$.

\subsection{Voting Mechanism}
As before mentioned, $I_1$, $I_2$, $I_3$ are three inference values predicted by our basic models, and a simple voting mechanism is applied to count the final output. When two or more models draw the same inference label, this label is our final prediction. The final output is predicted as:

\begin{equation}
f(x):=\left\{
\begin{aligned}
0&   \text{ if $I_1$+$I_2$+$I_3$ \textless  2 }\\
1&   \text{ if $I_1$+$I_2$+$I_3$ \textgreater= 2}
\end{aligned}
\right.
\end{equation}

\begin{figure}
\includegraphics[width=\textwidth]{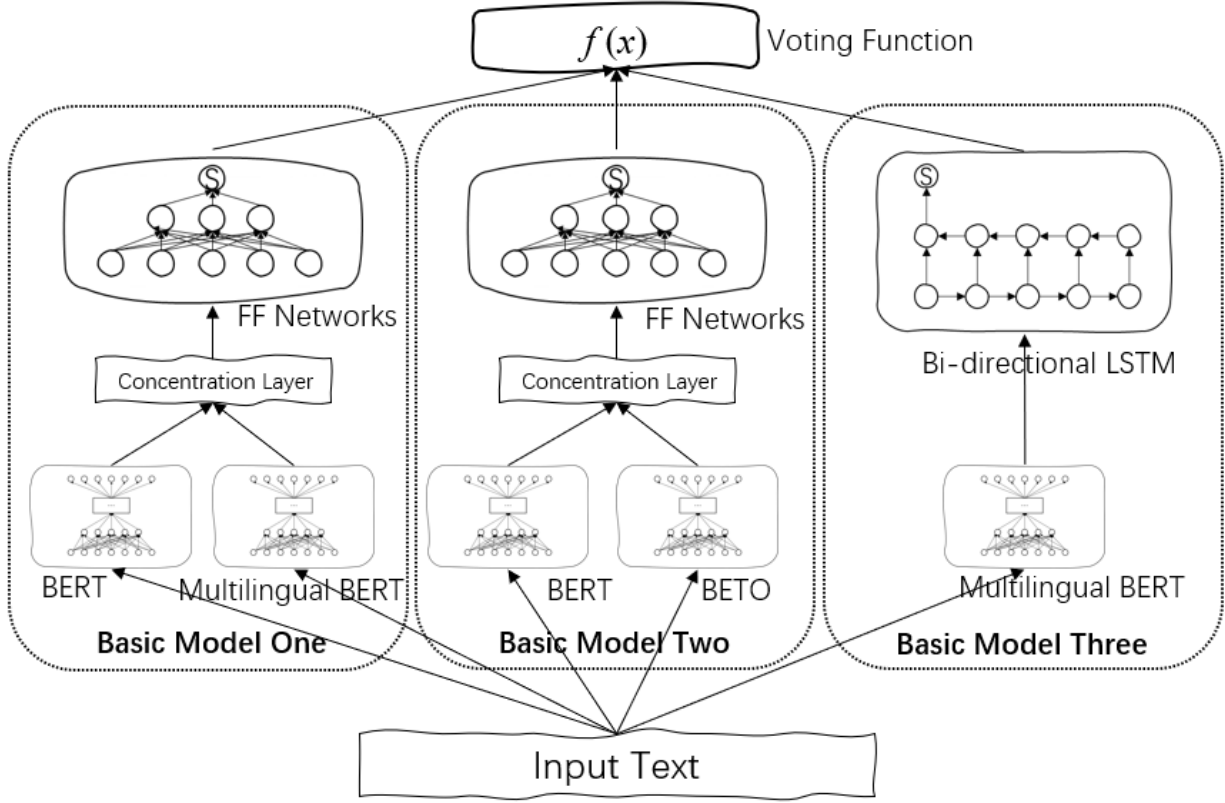}
\caption{The final model combines three basic models with a voting mechanism} \label{fig1}
\end{figure}

Our model architecture is shown in Figure \ref{fig1}, as we can see the input text date feeds into different basic model, and then flows to each inference network respectively, and finally all of these information are combined in the voting function for the prediction.

\section{Experiments and Results}

In this section, we present the description of dataset, the setup of our experiments, and the final results of our experiments. Experimental validation is conducted on the EXIST 2021 training and test corpus. The detaset is described in subsetion 3.1 followed by the analysis of our final results in  subsetion 3.2.

\subsection{Data Description}
There are two subtasks in this shared task, the first is Sexism Identification, which is a binary classification task, and the system needs to decide whether a given social media text is or is not sexist. The second subtask is to categorize these sexist texts into different predefined sexism types. This paper is focus on the first subtask.

Both of the training and test datasets consist in English and Spanish text data, which are collected from the social media such as twitter and grab. Table \ref{tab1} briefly describes the corpus. As we can see, there are 6977 sentences in the training set, and 4368 sentences in the test set, and labels are balanced in this subtask.

\begin{table}
\caption{Sentence numbers for each data set for subtask one.}\label{tab1}
\begin{center}
\begin{tabular}{|l|l|l|l|}
\hline
Date sets &  Language &  Label & Number of Sentence\\
\hline
  & en & non-sexist & 1800\\

Training & en & sexist & 1636\\

  & es & non-sexist & 1800\\

  & es & sexist & 1741\\
\hline
  & en &  non-sexist & 1050\\

Test & en &  sexist & 1158\\

  & es & non-sexist & 1037\\

 & es &  sexist & 1123\\
\hline
\end{tabular}
\end{center}
\end{table}

\subsection{Results and Discussion}
The evaluation of this sexism identification subtask is mainly based on Accuracy score. Our results are gathered in Table \ref{tab2}, which contains the Accuracy and F-Measure score for the basic models and the final systems. Our first question is that whether does the voting function bring additional benefits for our system. As we can see, the our final system (with a simple voting mechanism) performs the best Accuracy and F-measure metrics among these four systems, which provides a 3\% extra performance with the basic models.

\begin{table}
\caption{Evaluation Results: Accuracy and F-Measure}\label{tab2}
\begin{center}
\begin{tabular}{|l|l|l|l|}
\hline
Date set &  Model &  Accuracy & F1\\
\hline
  & Basic Model One & 0,7370 & 0,7360\\

  & Basic Model Two &  0,7280 & 0,7275\\

 Test & Basic Model Three & 0,7287 & 0,7287\\

  & Final Model (with voting mechanism) & \textbf{0,7553} & \textbf{0,7551}\\
\hline
\end{tabular}
\end{center}
\end{table}

Unlike the training data set, the test set contains the text data from different source, which are the twitter and gab. It brings the second question, that is whether our final model is data source sensitive. So we list the results in different data source for each model. Shown in Table \ref{tab3}, even the training data comes from twitter, but the results in gab is better than twitter data in the test set, which shows a robust ability of the language models. 

\begin{table}
\caption{Evaluation Results by Data Source: Accuracy}\label{tab3}
\begin{center}
\begin{tabular}{|l|l|l|l|}
\hline
Date set & Data Source &  Model  & Accuracy\\
\hline
& twitter & Basic Model One & 0,7271  \\
&  & Basic Model Two  &  0,7312\\
&  & Basic Model Three  & 0,7312 \\
Test &  & Final Model (with voting mechanism)& \textbf{0,7504 }\\
& gab & Basic Model One &  0,7709 \\
& & Basic Model Two &  0,7169 \\
&  & Basic Model Three & 0,7200 \\
&  & Final Model (with voting mechanism) & \textbf{0,7719 }\\
\hline
\end{tabular}
\end{center}
\end{table}

The last question about our system is whether our system is capable of processing multilingual data. Table \ref{tab4} lists the evaluation metrics of our systems by different languages. Results show that the performance of our models in different languages is similar to each other, and it seems that to introduce the language specific language models, i.e. the BETO has not brought extra benefits for our final system.  

\begin{table}[h]
\caption{Evaluation Results by Language: Accuracy}\label{tab4}
\begin{center}
\begin{tabular}{|l|l|l|l|}
\hline
Date set & Language &  Model  & Accuracy\\
\hline
& en & Basic Model One & 0,7197 \\
&  & Basic Model Two  &  0,7351 \\
&  & Basic Model Three  & 0,7486 \\
Test &  & Final Model (with voting mechanism)& \textbf{0,7559 }\\
& es & Basic Model One & \textbf{0,7546 }\\
& & Basic Model Two &  0,7208 \\
&  & Basic Model Three & 0,7083 \\
&  & Final Model (with voting mechanism) & \textbf{0,7546 }\\
\hline
\end{tabular}
\end{center}
\end{table}

\section{Conclusions}
In this paper, we implement a classification system with combining the language models and a simple voting mechanism. Our system has been evaluated on the EXIST 2021 subtask one. The evaluation results has shown that our system is data-source and language robust, and the voting function indeed brings in extra benefits to our final system. Our experiments prove that the pre-trained language model is also highly adaptable to social media texts, and a simple voting mechanism can highly leverage the predictive ability of the multi-model system.

\bibliographystyle{unsrt}

\end{document}